\pdfoutput=1

\documentclass[11pt]{article}

\usepackage{ACL2023}

\usepackage{times}
\usepackage{latexsym}

\usepackage[T1]{fontenc}

\usepackage[utf8]{inputenc}

\usepackage{microtype}

\usepackage{microtype}
\usepackage{booktabs}
\usepackage{hyperref}
\usepackage{url}
\usepackage{amsmath}
\usepackage{graphicx}
\usepackage{subcaption}

\newcommand\blfootnote[1]{%
  \begingroup
  \renewcommand\thefootnote{}\footnote{#1}%
  \addtocounter{footnote}{-1}%
  \endgroup
}

\title{FaceChat: An Emotion-Aware Face-to-face Dialogue Framework}

\author{
    Deema Alnuhait*,
    Qingyang Wu*, 
    Zhou Yu  \\
     Columbia University \\
    \texttt{\{daa2182,qw2345,zy2461\}@columbia.edu} \\
}

\begin{document}
\maketitle
\begin{abstract}

While current dialogue systems like ChatGPT have made significant advancements in text-based interactions, they often overlook the potential of other modalities in enhancing the overall user experience.
We present FaceChat, a web-based dialogue framework that enables emotionally-sensitive and face-to-face conversations. 
By seamlessly integrating cutting-edge technologies in natural language processing, computer vision, and speech processing, FaceChat delivers a highly immersive and engaging user experience.
FaceChat framework has a wide range of potential applications, including counseling, emotional support, and personalized customer service.
The system is designed to be simple and flexible as a platform for future researchers to advance the field of multimodal dialogue systems.
The code is publicly available at \url{https://github.com/qywu/FaceChat}.
\end{abstract}

\blfootnote{* Equal contribution.}

\section{Introduction}

The recent advancements in ChatGPT technology \cite{OpenAI_ChatGPT} have elevated the capabilities of dialogue systems to unprecedented levels of artificial intelligence.
Despite this progress, the most advanced models, such as ChatGPT are only optimized for text-to-text interactions, limiting their ability to provide a natural and immersive experience for the user. 
In contrast, early spoken dialogue systems \cite{10.3115/980691.980700,Dimitris_SDS} were designed to facilitate spoken interactions, offering a more human-like and engaging experience for the user.

Spoken dialogue systems offer several advantages over text-based chat interactions. 
It is a more natural and intuitive mode of communication for many people, and it can make conversation more engaging and interactive.
Also, for individuals who have difficulty typing or reading text, spoken dialogue systems can provide a more accessible and usable interface. 
It offers a more relaxed and casual atmosphere compared to typing-based interactions, creating a more comfortable and effortless environment for the user to engage with the system.

Spoken dialogue systems can be combined with multimodal technologies to create a face-to-face dialogue framework, often represented as digital humans in previous research. This integration can provide a more complete understanding of the user's emotional state through the capture of nonverbal cues such as facial expressions and tone of voice, leading to a more human-like and engaging interaction with real-time feedback.

\begin{figure}[t]
    \centering
    \includegraphics[width=0.47\textwidth]{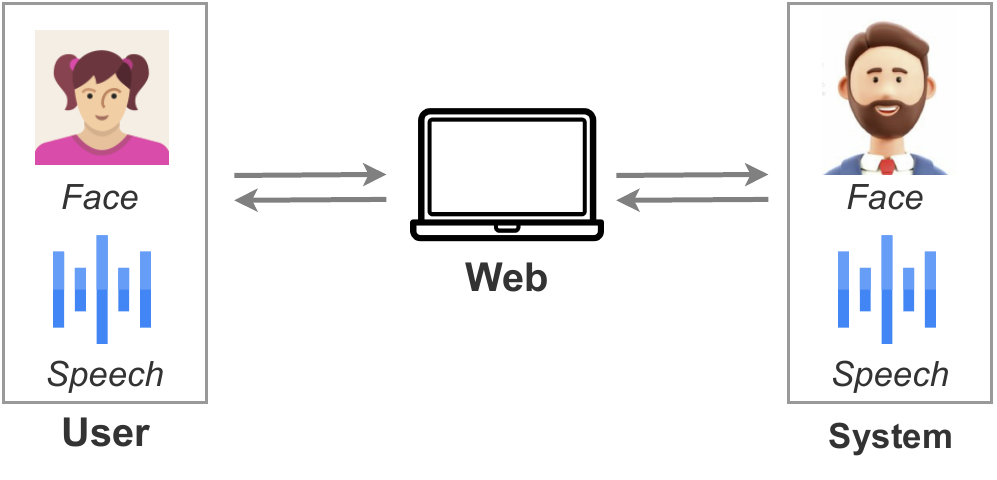}
    \caption{FaceChat delivers a seamless, real-time, and face-to-face dialogue experience through a web browser.}
    \label{fig:my_label}
\end{figure}



We present FaceChat, a cutting-edge chatbot designed to provide a seamless and immersive face-to-face dialogue experience via a web browser.
The development of emotion-aware face-to-face dialogue systems faces several challenges, including natural language understanding, interpretation of nonverbal cues, and real-time processing. 
To overcome these challenges, the pipeline has been optimized using WebRTC, and the models are executed on a GPU-based server to provide a more efficient and seamless user experience.

Our prototype features latest technologies, including WebRTC for fast data transmission, a pre-trained voice activity detector \cite{Silero_VAD} for monitoring user engagement, a neural network-based face detection model \cite{DBLP:conf/eccv/LiuAESRFB16} for accurate emotion detection, a GPT-3-powered chatbot \cite{NEURIPS2020_1457c0d6} with customizable character, VITS \cite{DBLP:conf/icml/KimKS21} for natural text-to-speech synthesis, and OpenAI's whisper for precise automatic speech recognition.
FaceChat delivers an emotionally-sensitive and interactive dialogue experience featuring real-time, face-to-face communication capabilities.
Furthermore, the backend is completely built on Python, which is easily integrable with popular deep learning frameworks to collect video and speech data for future advancements in multimodal dialogue systems.

To facilitate the development of a more advanced multimodal dialogue system, we open-source all our implementations.
Moreover, the majority of the services have been optimized for execution on a GPU-based server, eliminating the need for additional support systems.
The code is designed to run in the cloud, providing users with the ability to access the service from any location with a web browser, offering greater flexibility and convenience.
The code is publicly available at \url{https://github.com/qywu/FaceChat}.

\section{Related Work}
The development of dialogue systems has a rich, and extensive history \cite{chen2017survey}. 
Among the various types of dialogue systems, open-ended systems, also known as chit-chats, have received substantial attention in the field for their ability to facilitate general and non-task-specific conversations \cite{Serban2015BuildingED, jurafsky2018dialog}.

Dialogue systems can be text-based or spoken-based.
Spoken dialogue systems (SDS), as defined by \citet{article}, are computer systems that enable turn-by-turn conversations through speech.
A number of SDS frameworks have been proposed, with spoken language as the primary mode of interaction \cite{aicher-etal-2022-towards, 9889357}.

Multimodal dialogue systems build upon the capabilities of spoken dialogue systems (SDS) by incorporating visual modalities such as images and videos \cite{rudnicky2005multimodal}, thereby offering users a more immersive and engaging dialogue experience.
Multimodal HALEF \cite{zhouyu_multimodal_halef} and TickTock \cite{Yu2015TickTockAN} are examples of multimodal dialogue systems that improve response quality by incorporating user engagement detection.
Other works have explored projecting dialogue systems into the multimodality space by incorporating visual context such as images, and videos \cite{okur-etal-2022-data, electronics11203409, gan-etal-2019-multi}.
Despite these efforts, previous works did not integrate the newer deep learning models, and their performance is limited in terms of the current standards.

Emotion-aware dialogue systems are another area of research in the field. Previous studies have explored ways to enhance response quality by taking into account the user's emotions \cite{shi-yu-2018-sentiment, 9165162}. 
However, these works have been limited in their ability to accurately detect the user's emotions without incorporating multimodal information. 
In our approach, we achieve emotional awareness by integrating the user's facial expressions into the response generation process.








\begin{figure*}[t]
    \centering
    \includegraphics[width=0.85\textwidth]{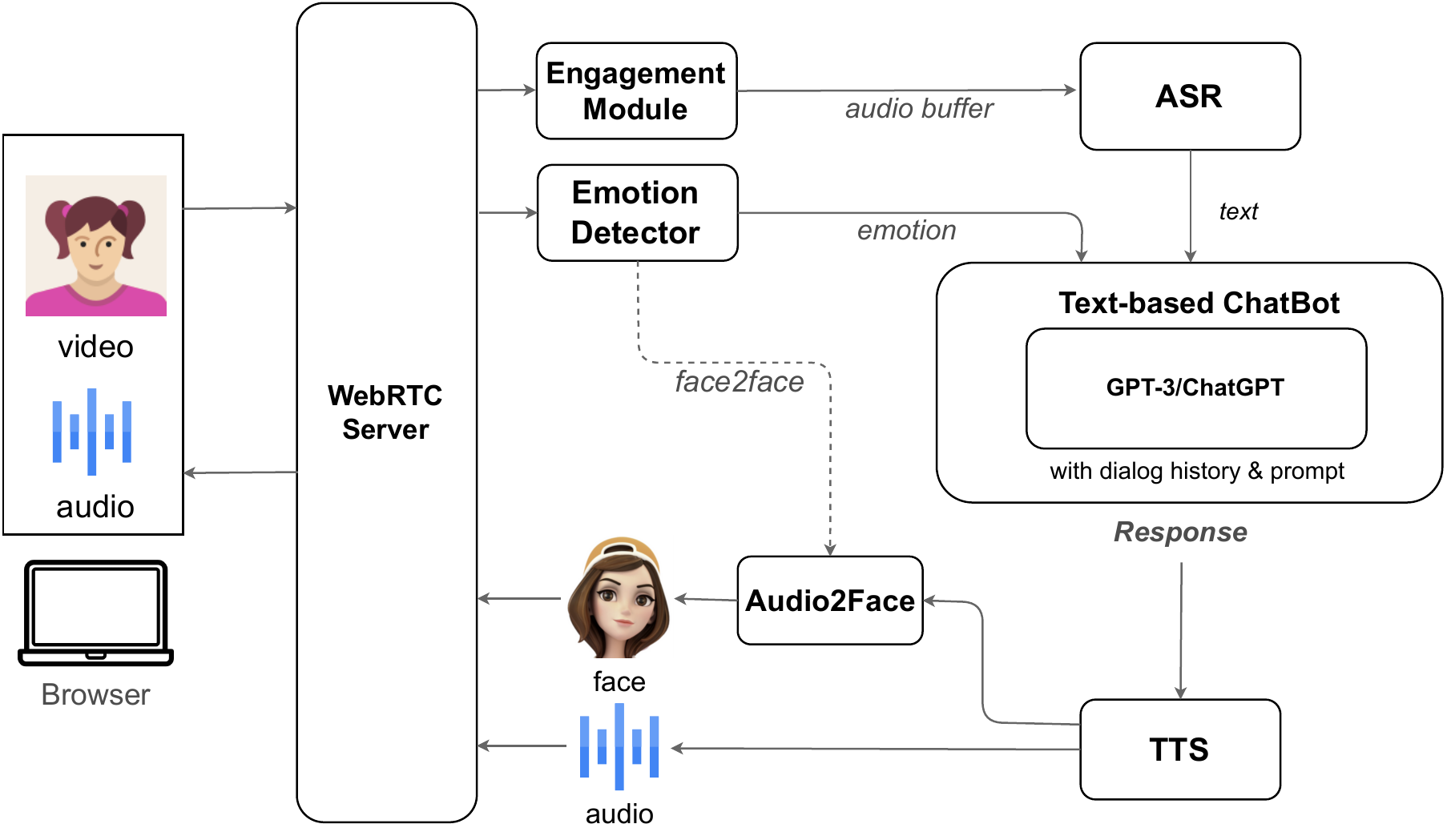}
    \caption{FaceChat Framework Architecture.  WebRTC server handles the video and audio data transmission between the user and the server. The backend is implemented with Python. }
    \label{fig:framework}
\end{figure*}

\section{FaceChat Framework}


FaceChat is a web-based face-to-face and emotion-aware dialogue framework that integrates several advanced modules, including a web module, an engagement module, an ASR module, a chatbot module, a TTS module, and a face generation module.
Figure~\ref{fig:framework} demonstrates the entire architecture of the FaceChat framework.
In the following sections, we will delve into the details of each module.

\begin{figure}[t]
    \centering
    \includegraphics[width=0.45\textwidth]{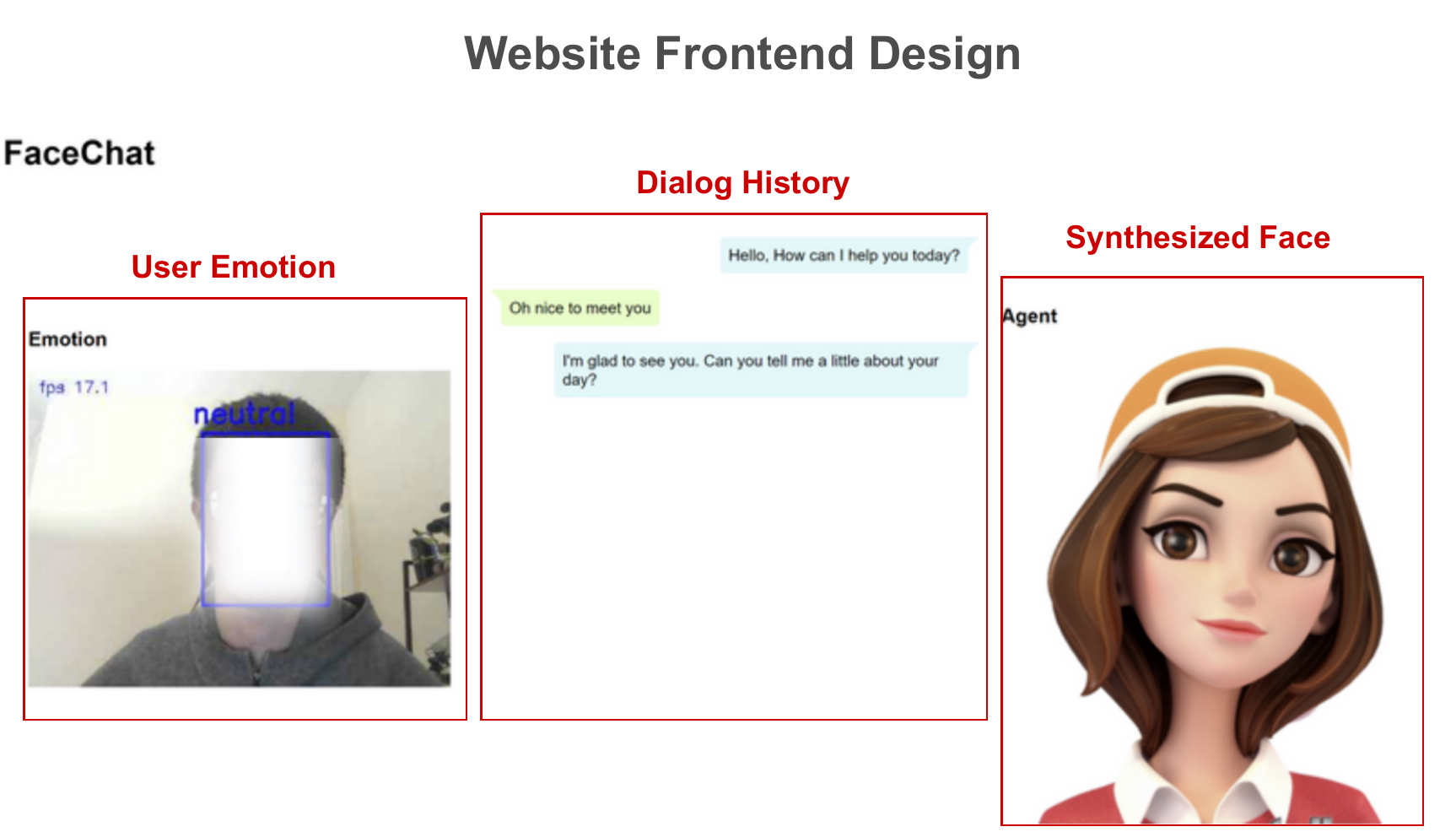}
    \caption{Frontend design of the interface}
    \label{fig:frontend}
\end{figure}

\subsection{Web-based System}

By having a web-based system, FaceChat eliminates the need for any additional setup, providing convenient in-browser access for all users.
We leverage Flask, a Python-based framework \cite{grinberg2018flask}, to build our application. 
Thus, FaceChat is easily integrable with popular deep learning packages such as PyTorch \cite{NEURIPS2019_9015} and TensorFlow \cite{abadi2016tensorflow}.
The system comprises two main components:
the frontend and the backend. The frontend focuses on the user-facing interface, and the backend deals with the underlying functionality and processing of the system.

\subsubsection{Frontend}
The frontend uses WebRTC \footnote{https://webrtc.org} to enable direct access to video and audio data from the user's camera and microphone through the web browser.
Then, raw video and audio data are streamed to the server in real-time via Socket.IO \footnote{https://socket.io/}.
To minimize latency, the microphone audio is downsampled from its original 44.1kHz to 16kHz.
Audio data is transmitted to the server in packets, each containing about 46.4ms audio segment (2048 block size).
The video data is captured at a resolution of 400x300 and a frame rate of 20 FPS.
The frontend is also responsible for playing the audio and image data transmitted back from the server, which achieves the functionality of face-to-face conversations.

The user interface of the demo system is depicted in Figure~\ref{fig:frontend}. 
The website is divided into three main sections, each serving a specific purpose. 
The first section focuses on the detection of the user's face and corresponding emotion. 
The second section provides a comprehensive dialog history, allowing the user to correct any previously misrecognized speech and providing a visual reference to the conversation. 
The final section showcases the synthesized face, incorporating basic lip-sync functionality to enhance the naturalness and engagement of the conversation. 
These three sections work together to create an immersive and interactive experience for the user.

\subsubsection{Backend}
The backend leverages eventlet and Flask-SocketIO \footnote{https://flask-socketio.readthedocs.io/en/latest/} for efficient, asynchronous event handling. 
Incoming video and audio packets are received and stored in a buffer for processing.
The engagement and emotion modules run in separate threads, asynchronously analyzing data from the buffer.
The engagement module determines the appropriate moments for triggering a response, while the emotion module captures the user's emotional state as input for the chatbot.
Subsequently, the ASR model and the text-based chatbot model will leverage the information obtained to generate a coherent and meaningful textual response.
Finally, the text-to-speech and audio-to-face modules will utilize the generated text and audio to create a dynamic and realistic representation of the face and sound, which will then be transmitted back to the user for a complete and immersive experience.

With a focus on ease-of-use and research efficiency, we put most of the data processing and computation on the server-side, freeing researchers from the complexities of frontend development.

\subsection{Engagement Module}

To determine user engagement, we utilize voice activity detection (VAD), with a two-stage approach combining the WebRTCVAD \footnote{https://github.com/wiseman/py-webrtcvad} and a model-based Silero VAD \cite{Silero_VAD} for improved precision and recall.

WebRTCVAD, a feature of the WebRTC library, enables real-time speech detection in audio streams through its Gaussian Mixture model. 
However, its sensitivity to human speech frequencies may result in many false positives. 
To mitigate this, we also employ Silero VAD, a pretrained model that excels in detecting speech from various backgrounds. 
Silero VAD requires sufficient context audio, and using it alone may result in false negatives when segmenting the audio. 
Thus, by combining both models, it achieves improved precision and recall.

\subsection{Emotion Classification}

The facial expression module operates in a separate thread, continuously monitoring the user's emotional state through facial expression recognition.
FaceChat employs DeepFace \cite{serengil2021lightface}, a Python-based library incorporating OpenCV \cite{opencv_library} and state-of-the-art face detection algorithms to process images in the buffer. 
For facial expression recognition, we use the fast and real-time single-stage object detection model SSD \cite{DBLP:conf/eccv/LiuAESRFB16} as the backbone.
Currently, FaceChat detects four emotions from the user: happy, sad, angry, and neutral.
The classified emotion is stored in a buffer to adjust the chatbot's response strategies.

\subsection{Automatic Speech Recognition}

Automatic speech recognition (ASR) is crucial for FaceChat's chatbot to process speech by converting it into text.
We use the latest ASR model, OpenAI's Whisper \cite{radford2022whisper}, a Transformer-based model trained on 680k hours of speech data. 
Whisper offers general-purpose speech recognition and supports multiple languages, performing speech translation and language identification simultaneously. 
Its voice activity detection function further reduces false positive speech detection.
In the FaceChat framework, Whisper is triggered by and processes the audio segments identified by the engagement module.
The recognized text is then passed on to the chatbot for the next state.

\subsection{Text-to-speech Synthesis}

To enable users to chat with our system using their voice, we use a technique called text-to-speech synthesis (TTS). With TTS, the system will convert the text responses from the chatbot into audio responses.
There are many TTS models available, and after careful consideration, we have chosen VITS \cite{DBLP:conf/icml/KimKS21} due to its fast processing time and ability to generate more natural-sounding speech than previous models.
We have integrated VITS using the publicly available code and pretrained model weights on the VCTK dataset \cite{Veaux2017CSTRVC} within the TTS package \footnote{https://github.com/coqui-ai/TTS}.
The synthesized audio is then played on the user's device.

\subsection{Talking Face Animation}

For our current talking head implementation, we use an external JavaScript library\footnote{https://github.com/talkr-app/talkr-apng} to synchronize the playback of APNG files with the generated audio.
However, the generated animations may appear unnatural, particularly in regard to lip-syncing, eyebrow, and blinking synchronization. 
For future work, we will integrate more powerful 3D talking avatars like Nvidia Audio2Face  \cite{10.1145/3072959.3073658}.

\subsection{Emotional Adaptive Dialogue Systems}

\begin{figure}[t]
    \centering
    \includegraphics[width=0.43\textwidth]{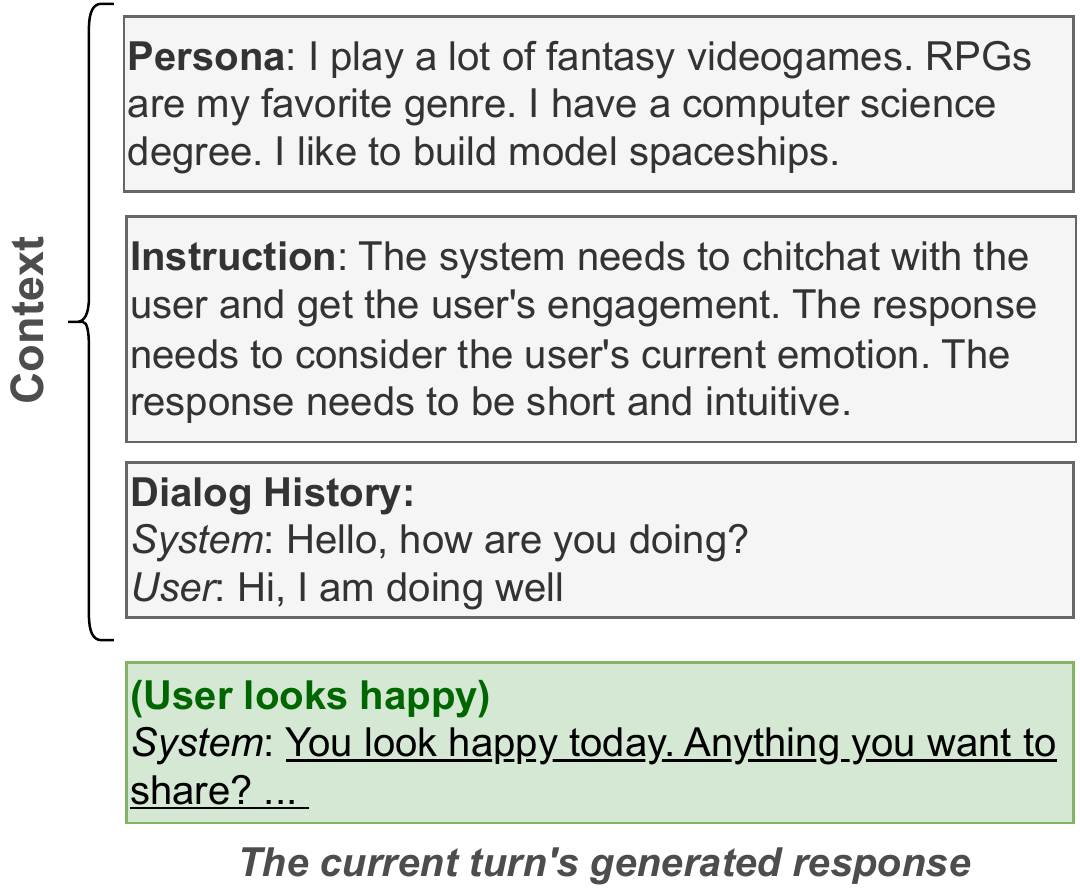}
    \caption{An example prompt to create a chatbot that integrates persona and emotion detection. }
    \label{fig:template}
\end{figure}

\begin{figure}[t]
    \centering
    \includegraphics[width=0.48\textwidth]{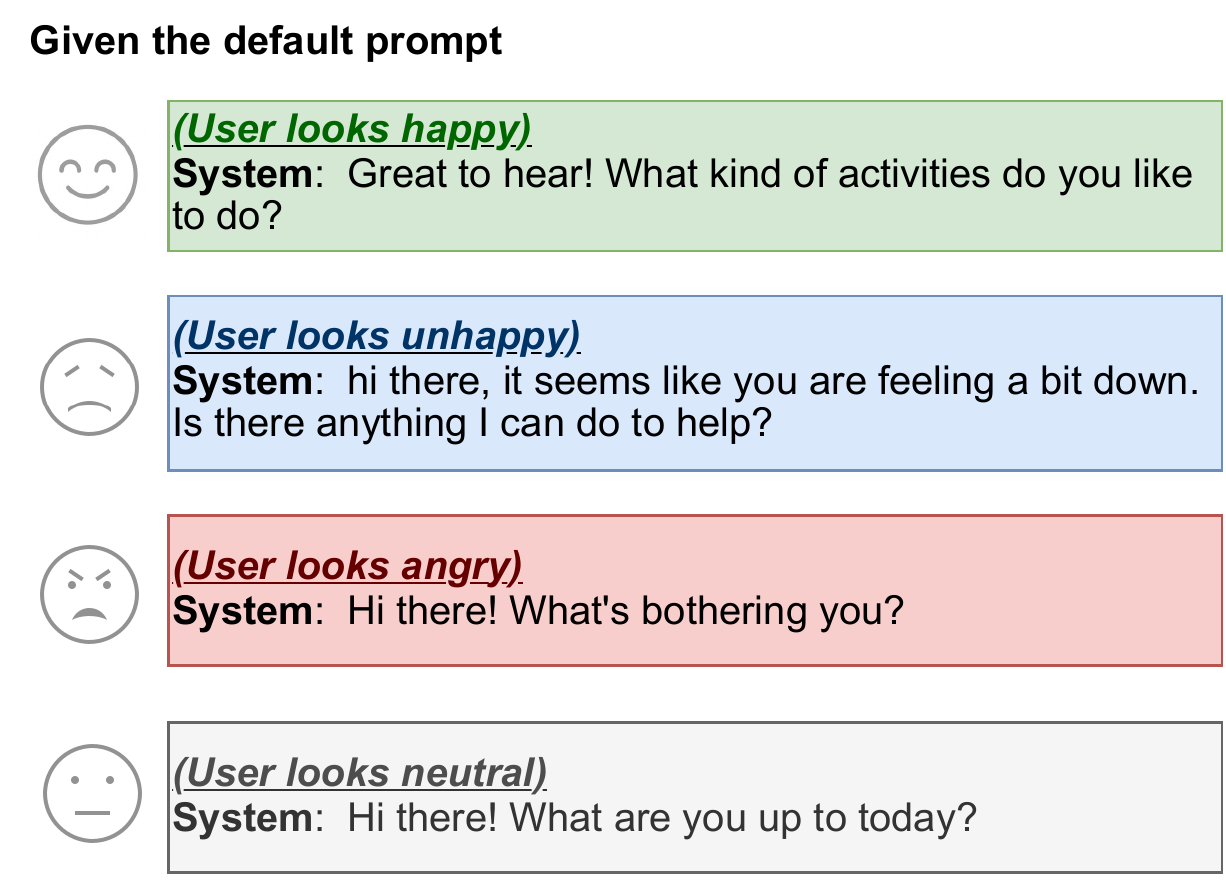}
    \caption{Based on different user facial emotions, the system can respond correspondingly.}
    \label{fig:emotion_response}
\end{figure}


The text-based dialogue system is a crucial component of our framework. Given the impressive capabilities of modern language models, we have opted to use a text-to-text dialogue system for its ability to provide strong zero-shot capabilities in both language generation and understanding. 
Our implementation offers an abstract interface for text-based chatbots,
making it possible to use other models in the future, and the system can be updated to incorporate new advancements in the field. 
Also, the chatbot is designed to interact with the user in real-time and is capable of generating natural and context-aware responses.

\subsubsection{Prompt for Emotion-Aware Chatbot}

GPT-3 \cite{NEURIPS2020_1457c0d6} is a state-of-the-art large language model developed by OpenAI. 
It has shown remarkable power in generating text and has been applied in many natural language processing tasks. 
In this work, we leverage its zero-shot capabilities to create an emotion adaptive chatbot for FaceChat.
One of the strengths of GPT-3 is that it can be customized to create different characters with different personalities, knowledge, and styles. This is achieved through the use of proper prompting techniques that provide the model with context and a clear definition of the desired character.
Thus, FaceChat can be extended to different applications, making it highly flexible.
Figure~\ref{fig:template} displays the prompt used for the emotion-aware chatbot, which consists of four key components. 

The first component is the persona of the AI agent, which draws inspiration from the PersonaChat dataset \cite{DBLP:journals/corr/abs-1801-07243}. 
The persona component of the agent serves to establish a basic character for the chatbot. 
It not only piques user interest and enhances engagement but also provides a natural starting point for the conversation.

The second component is the instruction component, which guides GPT-3 to behave like a chitchat agent, enabling it to generate appropriate responses that match the conversational context. 
In addition to considering the conversational context, the instructions guide the model to account for the emotional state of the user and adjust its responses accordingly.

The third component is the dialog history, which stores previous conversations in memory. 
By retaining this information, the system can maintain coherence and avoid generating responses that are out of context.

The final component generates the chatbot's response to the user's input. 
To account for the user's emotional state, we transform their facial expressions into a textual representation (e.g., the user looks happy). 
The model then utilizes this information to generate a response that is sensitive to the user's emotional state.

\begin{table*}[t]
    \centering
    \begin{tabular}{c|c c c c c}
        \toprule
        GPT \textbackslash Whisper & tiny & base & small & medium & large \\
        \midrule
        davinci & $1.76 \pm 0.29$ & $1.69 \pm 0.20$ & $1.74 \pm 0.39$ & $1.61 \pm 0.23$ & $2.03 \pm 0.23$ \\
        curie& $0.48 \pm 0.02$ & $0.65 \pm 0.06$ & $0.67 \pm 0.03$ & $0.67 \pm 0.03$ & $0.80 \pm 0.12$ \\
        babbage & $0.53 \pm 0.03$ & $0.39 \pm 0.02$ & $0.47 \pm 0.01$ & $0.66 \pm 0.14$ & $0.73 \pm 0.10$ \\
        ada & $0.40 \pm 0.06$ & $0.40 \pm 0.07$ & $0.47 \pm 0.02$ & $0.55 \pm 0.09$ & $0.66 \pm 0.09$ \\
        \bottomrule
    \end{tabular}
    \caption{Evaluation of latency performance in FaceChat using different GPT-3 and Whisper Models. 
    There is a trade-off between utilizing larger models and the potential increase in latency. 
    The latency is measured on the GPU RTX 3090Ti.
    }
    \label{tab:latency}
\end{table*}

\subsubsection{Analysis for Different Emotions}

The framework supports different emotions: happy, sad, angry, and neutral.
When a user interacts with the chatbot, the chatbot will use the facial expression to identify the user's current emotional state. 
The emotion state will then be used in the template prompt, as shown in Figure~\ref{fig:template}, to generate the current turn's system response.

To evaluate how the system response would vary when different emotions are expressed by users, we conducted tests where the same context was presented with varying emotional expressions. 
The goal of these tests was to analyze how the system responds to different emotions and how it can provide more appropriate and effective responses.

The results of our testing are presented in Figure~\ref{fig:emotion_response}.
When the user expresses happiness, the system responds with positive messages that help to create a positive and friendly conversation.
When the user expresses sadness, the system responds in a caring and empathetic manner. 
It tries to provide comfort and support by offering help or guidance.
When the user expresses anger, the system responds in a calm and measured manner, with the goal of de-escalating the situation and understanding the user's concerns more fully. 
When a user expresses a neutral emotional state, the system recognizes that the user may not be fully engaged or invested in the conversation. 
In this situation, the system may try to engage the user by asking questions or providing prompts that are designed to pique the user's interest.



\section{Latency Evaluation}
Ensuring low latency is a critical factor for our system as it directly impacts the user experience. 
Therefore, we conduct thorough evaluations to measure the latency of our entire system. 
We measure the latency between when the user finishes speaking and when the system starts to speak.
We find that the main sources of latency come from the ASR model Whisper and the GPT-3-based chatbot. 
As there is a trade-off between model sizes and latency, we evaluate the performance of various combinations of GPT-3 models and Whisper models to optimize the system's response time. 

The results are shown in Table~\ref{tab:latency}.
The utilization of GPT APIs may result in a noticeable lag.
The Whisper model is running locally, and thus it takes a relatively smaller amount of latency.
We can observe that using the largest GPT model, davinci, would result in a delay of more than one second.
As a trade-off, it provides the best dialogue quality.

Our evaluations have found that users typically find a response delay of approximately one second to be acceptable
while it is noticeable when the delay exceeds two seconds.
Based on these findings, we have decided to use a combination of GPT-3 davinci model and the Whisper medium model in our final system. 
This allows us to balance between model complexity and response time, achieving overall a better user experience.

\section{Future Work}

There is much room for future advancements in the current system including,
ChatGPT integration to enhance the chatbot's capabilities 
and incorporating additional modalities such as voice emotion or engagement to improve the dialogue experience.
Although the current TTS technology has achieved considerable success, it continues to face challenges in generating speech with appropriate emotion and intonation. 
Particularly in a conversational context, the lack of contextual awareness often leads to an unnatural tone in the generated speech.
To address this limitation, we plan to investigate more advanced and controllable text-to-speech systems, such as VALL-E \cite{https://doi.org/10.48550/arxiv.2301.02111}.

To furtherly improve the facial animation, 
we aim to investigate advanced blend shape generation techniques, similar to the one employed by Nvidia's Audio2Face \cite{10.1145/3072959.3073658}, to enhance the realism of the facial animations. 

Our current framework does not support interruption from the user.
To achieve this, the system must precisely identify the user's intention to interrupt, 
as well as distinguish its own speech from the user's speech to prevent any interference.
When successfully interrupted by the user, the system should also adjust its response in a seamless manner to maintain the natural flow of the conversation.
We anticipate that a jointly trained VAD, ASR, and TTS model will bring about this capability.

\section{Conclusion}

In this work, we present FaceChat, a dialogue framework that incorporates the latest technologies and offers users a seamless and immersive face-to-face dialogue experience with emotional awareness.
FaceChat can be utilized in a variety of applications that demand face-to-face dialogues to accurately gauge the user's emotional state and engagement, such as customer service, counseling, and emotional support.
FaceChat has been engineered for ease of use, with a Python-based backend that facilitates easy integration with other algorithms and models.
This framework is well-suited for future research and advancements in the field of multimodal dialogue systems





\section*{Ethical Consideration}
The current design of FaceChat integrates large language models which have been trained on internet data. This made the entire pipeline prone to generate nonfactual, contradicted, or hateful content.
Moreover, FaceChat is susceptible to malicious use including data scrapping, fraud, or intruding on the user's privacy.
In addition to negatively affecting the user's emotions if it detects the emotions incorrectly.

\section*{Acknowledgement}

We would like to thank the Sony Research Program for supporting this project and Fuminori Homma, Yuki Okamura, Aditya Vavre, and Tsunayuki Ohwa for the discussion.

\bibliography{custom}
\bibliographystyle{acl_natbib}




\end{document}